\documentclass[runningheads]{llncs}
\usepackage{graphicx}
\usepackage{comment}
\usepackage{amsmath,amssymb} % define this before the line numbering.
\usepackage{color}

\usepackage[width=122mm,left=12mm,paperwidth=146mm,height=193mm,top=12mm,paperheight=217mm]{geometry}
\usepackage{caption}
\usepackage{subfig}
\usepackage{balance}
\usepackage{booktabs}
\usepackage{multirow}

\usepackage{microtype}
\usepackage[pagebackref=true,breaklinks=true,letterpaper=true,colorlinks,bookmarks=false]{hyperref}

\begin{document}
% \renewcommand\thelinenumber{\color[rgb]{0.2,0.5,0.8}\normalfont\sffamily\scriptsize\arabic{linenumber}\color[rgb]{0,0,0}}
% \renewcommand\makeLineNumber {\hss\thelinenumber\ \hspace{6mm} \rlap{\hskip\textwidth\ \hspace{6.5mm}\thelinenumber}}
% \linenumbers
\pagestyle{headings}
\mainmatter
\def\ECCVSubNumber{1212}  % Insert your submission number here

\title{ Person-in-Context Synthesis \\ with Compositional Structural Space} % Replace with your title

% INITIAL SUBMISSION 
\begin{comment}
\titlerunning{ECCV-20 submission ID \ECCVSubNumber} 
\authorrunning{ECCV-20 submission ID \ECCVSubNumber} 
\author{Anonymous ECCV submission}
\institute{Paper ID \ECCVSubNumber}
\end{comment}
%******************

% CAMERA READY SUBMISSION
%\begin{comment}
\titlerunning{Abbreviated paper title}
% If the paper title is too long for the running head, you can set
% an abbreviated paper title here
%
\author{Weidong Yin\inst{1} \and
Ziwei Liu\inst{2} \and
Leonid Sigal\inst{1}}
%
% First names are abbreviated in the running head.
% If there are more than two authors, 'et al.' is used.
%
\institute{University of British Columbia \\
\email{\{wdyin,lsigal\}@cs.ubc.ca} \and
The Chinese University of Hong Kong \\
\email{zwliu.hust@gmail.com}}
%\end{comment}
%******************
\maketitle

\vspace{-5pt}
% \begin{center}
%     \centering
%     \includegraphics[width=1\linewidth]{figure/front_fig.pdf}
%     \captionof{figure}{An illustration of the difference between layout to image synthesis, pose guided synthesis and person in context synthesis. 
%     First column illustrates an example from~\cite{sun2019image}; 
%     %  At the first column an example from~\cite{sun2019image} is provided. 
%     the persons are blurry with body parts unclear. 
%     The second column illustrates result from~\cite{Park2019GauGANSI}.
%     % At the second column result from~\cite{Park2019GauGANSI} is provided.
%     The model only deals with one single pose as input and does not model context information. In third column and onward 
%     % Beginning from the third column, 
%     we illustrate our results. The generated persons have clear body parts, respect the input poses and are compatible with specified layout contexts (top).}
% \end{center}%

%\thispagestyle{empty}

%%%%%%%%% ABSTRACT
\begin{abstract}
   \vspace{-12pt}
Despite significant progress, controlled generation of complex images with interacting people remains difficult. Existing layout generation methods fall short of synthesizing realistic person instances; while pose-guided generation approaches focus on a single person and assume simple or known backgrounds. To tackle these limitations, we propose a new problem, \textbf{Persons in Context Synthesis}, which aims to synthesize diverse person instance(s) in consistent contexts, with user control over both. The context is specified by the bounding box object layout which lacks shape information, while pose of the person(s) by keypoints which are sparsely annotated. To handle the stark difference in input structures, we proposed two separate neural branches to attentively composite the respective (context/person) inputs into shared ``compositional structural space'', which encodes shape, location and appearance information for both context and person structures in a disentangled manner. This structural space is then decoded to the image space using multi-level feature modulation strategy, and learned in a self supervised manner from image collections and their corresponding inputs.
Extensive experiments on two large-scale datasets (COCO-Stuff \cite{caesar2018cvpr} and Visual Genome \cite{krishna2017visual}) demonstrate that our framework outperforms state-of-the-art methods w.r.t. synthesis quality.% Notably, our approach improves the performance ($\sim$30\% absolute gains in FID) on COCO-Stuff test sets at 256 resolution under person-in-context setting, and shows substantial advantage under the scenario of multiple persons as well as diverse human poses.  

%the appearance of both is modeled by per-instance latent code, inferred or drawn from a distribution.  

   % Existing scene/layout generation methods fall short in synthesizing 
   % person instances, while person generation works assume simple backgrounds and ignore synthesis of surrounding image context. To tackle these challenges, we propose a new problem, \textbf{Persons in Context Synthesis}, which aims, with user control, to synthesize 1) diverse person instances, as well as 2) varying context that is visually compatible with the synthesized person(s). Unlike previous works that either take layout or keypoint as input, the proposed problem specify both of them, whose modalities are completely different from each other. Thus these raw annotations are projected onto the ``compositional structural space'' using two separate neural branches, where person and context representations are compatible with each other. Then we design multi-level feature modulation strategy and person-context discriminator to learn this structural space in a weakly supervised manner. Extensive experiments on two large-scale datasets (COCO-Stuff and Visual Genome) demonstrate that our framework outperforms other state-of-the-art methods w.r.t synthesis quality and diversity. Notably, our approach drastically improves the performance ($\sim$30\% absolute gains in FID) on `person split' test set, and shows substantial advantage under the scenario of multiple persons as well as diverse human poses.  
\keywords{Image Synthesis and Manipulation; Pose Guided Image Synthesis; Generative Models}
   
\end{abstract}

%%%%%%%%% BODY TEXT

\begin{figure}[t]
\begin{center}
    \centering
    \includegraphics[width=1\linewidth]{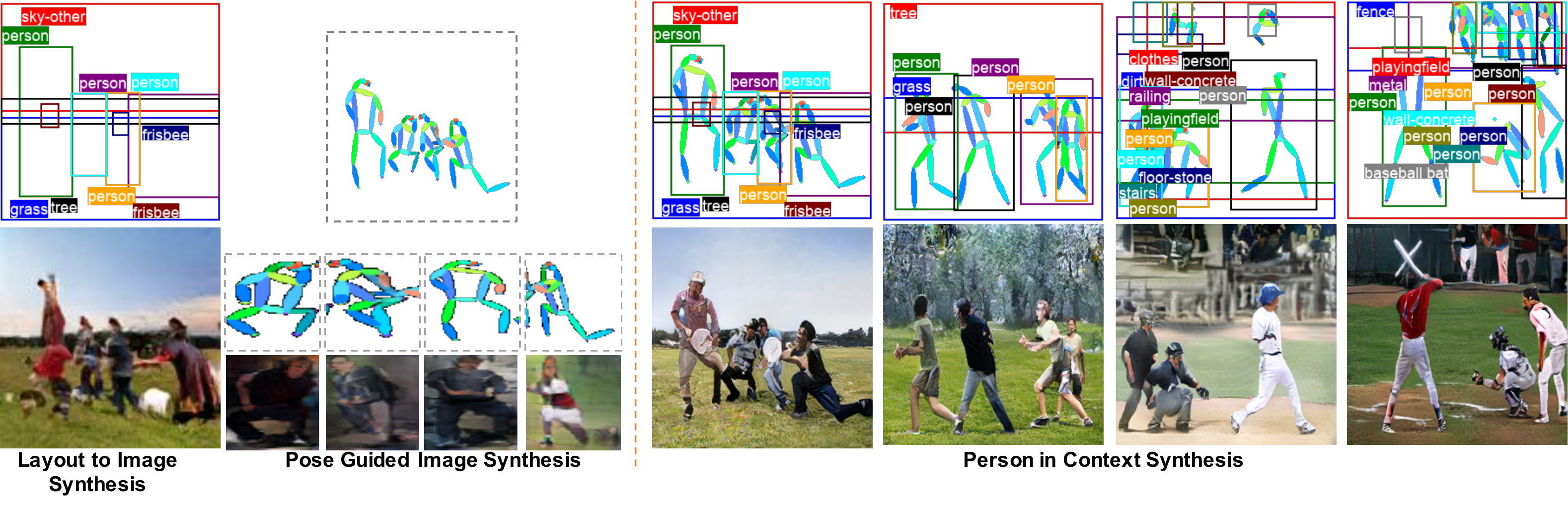}
    \captionof{figure}{{\bf Generative Settings.} An illustration of the difference between layout to image synthesis, pose guided synthesis and person in context synthesis (proposed). 
    First column illustrates an example from~\cite{sun2019image}; 
    %  At the first column an example from~\cite{sun2019image} is provided. 
    %the persons are blurry with body parts unclear. 
    The second column illustrates result from~\cite{Park2019GauGANSI}.
    % At the second column result from~\cite{Park2019GauGANSI} is provided.
    %The model only deals with one single pose as input and does not model context information. 
    In third column and onward 
    % Beginning from the third column, 
    we illustrate our results. 
    %The generated persons have clear body parts, respect the input poses and are compatible with specified layout contexts (top).
    }
    \label{fig:setting}
    \vspace{-30pt}
\end{center}%
\end{figure}

\vspace{-25pt}
\section{Introduction}

Learning to synthesize complex scenes with multiple persons and objects is one of the core problems in computer vision. Such technology may fundamentally revolutionize image search, as well as provide insights for visual inference problems.
Many recent works tackle the problem using layouts, which is a powerful structured representations for encoding the classes and locations of objects. For example, \cite{ashual2019specifying,johnson2018image} use layout as an intermediate representation between scene graphs and images. Alternatively, \cite{sun2019image,zhao2019image} directly take layout as input to generate images. While being able to generate limited objects with simple structures, existing works fail to model `person' faithfully, see Fig. 1-left. 
% It is due to the nature of its complex structures and diverse appearances. 
This observation is also supported in~\cite{bau2019seeing}, where GANs fail to reconstruct the person of the original image. Presumably the challenge is the diversity of human articulation and appearance. 

%  On the other hand, pose is a powerful guidance towards the synthesis of persons. 
%  Some recent works
In separate research thread, 
\cite{Balakrishnan2018SynthesizingIO,Han2017VITONAI,ma2017pose,Siarohin2017DeformableGF} focus on synthesizing persons with pose as a powerful guidance. Most of these works take raw image containing background as input. They then manipulate the original person(s) in that image towards provided pose(s). There are several drawbacks for these methods: 1) they do not model the context for corresponding person, thus either the background is simple or is provided as part of the input image. 2) they can only model one person per image, lacking the interactions between different person instances.

To overcome these limitations, we propose a new task called \textbf{Persons in Context Synthesis}, which aims to synthesize diverse person instance(s) in the specified layout context (see Figure~\ref{fig:setting} for illustration). 
% as well as varying contexts that are visually compatible with the synthesized person(s). 
By specifying the input layout and keypoints inside each person box, our approach is able to generate a high-resolution realistic image that contains the desired context and compatible person instance(s). In this manner, we jointly model the interactions between and among persons and objects, 
% different persons and different objects
within a unified framework.

Several unique challenges arise with this new task. First, layouts and keypoints are %annotations with 
fundamentally different modalities. Previous works only deal with single input modality. Naive combination of these two research streams does not yield satisfactory results.
% , as it is challenging to incorporate these two annotations into one single framework. 
Second, the information conveyed by layouts and keypoints is limited. Unlike semantic image generation tasks that leverage masks, the input here contains limited spatial information. The actual shape and appearance of object(s) and person(s) should be determined by not only the locations, labels and keypoints, but also their interactions and compatibility in the scene. % under the desired context. 
A good generative model should take all of these factors into consideration.

In this work, we address the above challenges by modeling layouts and keypoints using two separate neural branches, namely {\em context} and {\em person} branch respectively, which attentively composite the respective inputs into shared {\em compositional structural space}. This learned structural space is beneficial for final synthesis in many aspects. First, the shape, location and appearance of each person, or context object, is represented and encoded in a disentangled manner. Second, the person and context structures are compatible with each other and can be composited in this mid-level space with simple linear summation. 
%Third, compositional structural (latent) space can be learned directly from images with corresponding coarse annotations. 
Third, the compositional structural space can be learned in a self supervised manner from image collections and corresponding inputs, with proposed multi-level feature modulation strategy and person-context discriminator.
Finally, it enables high-quality and high-resolution image synthesis, and shows performance boost in FID on proposed `person split' test set. %In the context branch, each object is disentangled into two parts, namely label part and appearance part. Based on this representation we generate instance-level attention for each object construct feature map for each object with instance attention and object representation. Then the object feature map is warped into locations specified by corresponding bounding boxes, and then fused into \textbf{layout guided context canvas}. For the person branch, we crop out feature maps from context canvas at the location specified by person bounding boxes, which provides contextual information for the person to generate. Then we feed the cropped contextual map with pose heatmaps specified by keypoints as input into a U-Net, so that the network learns to convert it into person canvas compatible with the context canvas. Then for each person we use same warping to put them into correct locations, and fuse them up into \textbf{keypoint guided person canvas}. Then structure canvas is generation by fusion of person canvas and context canvas.
%Unlike previous works that directly take bounding boxes, keypoints or masks as input, our structural canvas lies in a latent space that attentively encodes both persons and context information in a location-aware manner. 

\vspace{0.05in}
\noindent
\textbf{Contributions.} Our contributions are three-fold: {\bf 1)} We propose a new task called persons in context synthesis, which takes both keypoints and layouts as input, and aims to synthesize diverse person instances as well as varying contexts that are visually compatible with the synthesized person(s). {\bf 2)} To handle the stark difference in input structures,we proposed two separate neural branches to attentively composite the respective (context/person) inputs into shared ``compositional structuralspace'', which encodes shape, location and appearance information for both context and person structures in a disentangled manner. 
%a novel framework called structural canvas, which lie in a latent space that attentively encodes both persons and context information in a location-aware manner.
%Above contributions, jointly, allow user controlled generation of high-resolution  diverse appearance images.
% which enables the possibility to synthesize images with photorealistic and diverse appearances. 
{\bf 3)} We performed extensive evaluations on two large-scale datasets (COCO-Stuff \cite{caesar2018cvpr} and Visual Genome \cite{krishna2017visual}) to demonstrate that our framework outperforms state-of-the-art methods in synthesis quality and diversity.

\vspace{-10pt}
\section{Related Work}

%\subsection{Conditional Image Generation Tasks}  
\noindent
\textbf{Conditional Image Generation.}
Conditional image generation approaches generate images conditioned on additional input information, including semantic maps  \cite{Isola2016ImagetoImageTW,Park2019GauGANSI,Wang2017HighResolutionIS}, image captions \cite{li2019object,yin2019semantics}, sketches \cite{Chen2018SketchyGANTD,Lu2017ImageGF} and input images \cite{Liu2017UnsupervisedIT,Zhu2017UnpairedIT,Zhu2017TowardMI}. %In our work, generating images from layout is also a specific kind of conditional image generation, where the input condition is given by layout consisting of bounding boxes, their corresponding class labels and keypoints for person(s). 
%To generate a diverse set of outputs, we introduced object embeddings as an additional condition so that the model can generate different style of outputs or specify the style for each bounding boxes conditioned on different object embeddings.
%\noindent
%\textbf{Image Generation from Layout.} 
%
Generating images from layout is also a specific kind of conditional image generation task. Layout is often used as an intermediate representation during the generation process, e.g., when generating from text \cite{li2019object} or scene graphs \cite{johnson2018image}. However, such approaches fail to generate images of high quality. 
% these constraints are too weak for the model to generate images with high quality. 
In contrast, \cite{johnson2018image} generate images from provided semantic map, 
%Some existing works generate images from provided semantic map \cite{johnson2018image},
achieving high quality results at the expense of very laborious pixel-level user input.
% which achieves high quality because given semantic map contains fine-grained shape information for each instance. 
Different from these works, we try to generate images directly from the given layout and keypoints, which is a novel and fundamentally different paradigm for image generation.

\vspace{0.05in}
\noindent
\textbf{Pose Guided Image Synthesis.} 
Recently, several GAN-based models  ~\cite{Balakrishnan2018SynthesizingIO,Han2017VITONAI,ma2017pose,Siarohin2017DeformableGF} have been proposed % to use adversarial training 
for pose guided image synthesis. Most of these works take raw image as input and generate images with different pose by borrowing information from the raw input image. In contrast, \cite{ma2018disentangled}
use sampling, in the disentangled latent space, to generate person images.
% applied disentangle learning to generate person images without any raw image provided by sampling from a latent space. 
However, these approaches  
% either they 
learn to predict a person in a new pose on top of the specified training background or even require empty, white background.
% , and do not model the context. 
Instead, our method models both complex background context and persons jointly in a unified framework.

\vspace{0.05in}
\noindent
\textbf{Feature Modulation Techniques.} 
%
%\noindent
%\textbf{Conditional Normalization.} 
%
Conditional normalization layers \cite{Huang2017ArbitraryST,Vries2017ModulatingEV} were first proposed in the task of style transfer, and then applied to other kinds of tasks. Most of these conditional normalization layers work by first normalizing the layer activations into zero mean and unit variance. Then they are denormalized into different mean and variance using learned affine transformaitons conditioned on external data such as class labels. % But these 
The earlier normalization techniques produce uniform normalization parameters across spatial locations, washing away class information across different spatial locations. For these reasons we adopt the spatial adaptive normalization layer~\cite{Park2019GauGANSI}. In our work, the normalization parameters are generated from compositional structural canvas to provide guidance towards final image synthesis. Thus we preserve the structural information during the generation process.

\vspace{0.05in}
\noindent
\textbf{Attention Mechanisms.}
Attention was first proposed in machine translation and then widely applied in various vision tasks such as classification \cite{Hu2017SqueezeandExcitationN,Wang2017ResidualAN}, image captioning \cite{Anderson2017BottomUpAT} and generative models \cite{Mejjati2018UnsupervisedAI,Zhang2018SelfAttentionGA}. Most attention mechanisms work by generating attention masks and then aggregating features with these provided masks. 
The resulting dynamic feature aggregation strategies enhance traditional neural networks.
% This mechanism provides a different kind of information interaction, thus enhancing the modeling ability of neural network. 
In this work, we proposed instance-level attention to better model the diverse shapes and varying appearances of different objects.
\begin{figure*}[t]
\begin{center}
   \includegraphics[width=1\linewidth]{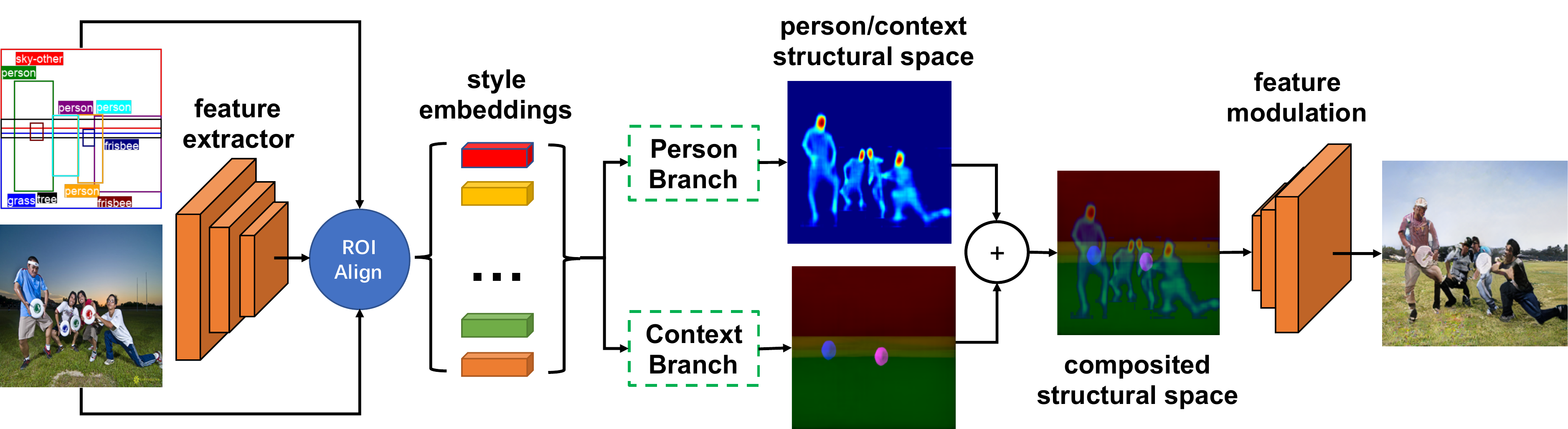}
\end{center}
   \vspace{-5pt}
   \caption{{\bf Overview of our framework.} The input to our model (in training) is the ground truth image with its layout and keypoints. High level feature maps are first extracted from ground truth using ResNet50. Then we use ROIAlign to crop out feature maps for different instances including persons and objects. Style embedding for each object is generated using VAE given the cropped feature map, then fed into person branch and context branch respectively.
   These two branches project layout and keypoint annotations into shared \textit{compositional structural space} conditioned on the style embeddings. Finally, we perform multi-level feature modulation to decode this structural space to a final image. }
   \vspace{-20pt}
\label{fig:overview}
\end{figure*}

\vspace{-20pt}
\section{Our Approach}

\newcommand{\norm}[1]{\left\lVert#1\right\rVert}

%  \noindent
%  \textbf{Framework Overview.}
%
Our goal is to develop a model which takes as input the context and person representations and synthesizes realistic image % of person in context 
correspondingly. The context is represented by layout consisting of bounding boxes and their class labels while person(s) are specified by keypoints in corresponding bounding boxes. The primary challenges are as follows: First, the layout as context representation is coarse and synthesized images need to respect the location of bounding boxes, class labels and style embeddings specified by the input. Second, the synthesized person instances need to be diverse and respect the pose(s). Finally, the synthesized image of person in context need to be compatible and realistic with natural interactions between and among person(s) and object(s).
%, different persons and different objects. 

To address these challenges, we introduce two key components in our framework, namely {\em person branch} and {\em context branch}. These two branches are used to model two different types of annotations separately and project them into the same \textit{compositional structural space}, which undergoes multi-level feature modulation in decoding to obtain a synthesized image. See  Figure~\ref{fig:overview} for illustration.
Notably, all components are differentiable % in a plug-and-play manner, 
and trained end-to-end without any extra supervision needed, except for the ground truth images with aforementioned annotations. 
We will introduce components in detail in following sections.

\vspace{-10pt}
\subsection{The Construction of Compositional Structural Space}
\vspace{-5pt}

\noindent
\textbf{Person in Context Layout.} 
The input to our model is person in context layout. It consists of two parts, namely context layout and multiple poses. During training, ground truth image is also needed. Specifically, 
%given the context layout $L$ with corresponding image $I_l$ and outputs a canvas $C_l=G_l(L,I_l)$. 
given a set of object categories $C$, a person in context layout $L$ is a tuple $(O,B,K)$ where $O=\{c_1,\dots,c_n\}$ is a set of objects with class types $c_i\in C$, and $B=\{\mathbf{b}_1,\dots,\mathbf{b}_n\}$ is a set of coordinates, $\mathbf{b}_i \in \mathbb{R}^4$, of the form $(x_1,y_1,x_2,y_2)$, where $(x_1,y_1),(x_2,y_2)$ is the upper left corner and lower right corner of the corresponding bounding boxes respectively. Bounding boxes are divided into two types, where $B_{o}=\{\mathbf{b}_{o1},\dots,\mathbf{b}_{on_o}\}$ do not contain person and $B_{p}=\{\mathbf{b}_{p1},\dots,\mathbf{b}_{pn_p}\}$ contain person; $n_o+n_p=n$ and $B = \{ B_{o}, B_{p} \}$. For each $\mathbf{b}_{pi}$ we have corresponding keypoints $K=\{\mathbf{k}_1,\dots,\mathbf{k}_{n_p}\}$, where $\mathbf{k}_i = \{ (\hat{x}_1, \hat{y}_1), (\hat{x}_2, \hat{y}_2), ..., (\hat{x}_m, \hat{y}_m) \}  \in \mathbb{R}^{2m}$. 

\vspace{0.05in}
\noindent
\textbf{Object Embeddings from RoIAlign.}
Given the ground truth image, we first extract feature map using ResNet50~\cite{he2016deep}. Then object embeddings corresponding to all bounding boxes, including person and context objects, are cropped using ROIAlign \cite{he2017mask} from the extracted feature map. The object embeddings $\mathbf{o}_i \in \mathbb{R}^{512}$ are used to model and control appearance (color/texture) of different objects. 

\vspace{0.05in}
\noindent
\textbf{Diverse Style Embeddings.}
%
% The extracted object embeddings does not follow gaussian prior. 
The extracted object embeddings, by default, do not follow any distribution that can be easily sampled at test time. 
To be able to sample diverse images with different styles of objects, we introduced a VAE \cite{kingma2013auto} which takes extracted object embeddings $\mathbf{o}_i$ as input and generate corresponding style embeddings $\mathbf{e}_{oi}$ by sampling from the posterior $Q(\cdot|\mathbf{o}_i)$. 
At test time we sample from Gaussian prior instead to get diverse appearances for both persons and objects.
% Thus we can sample from Gaussian distribution to get diverse appearances for both persons and contextual objects at test time. 
KL loss is introduced to regularize the network: 
% from overfitting on object embedding:
\begin{equation}
    \mathcal{L}_{KL}=\mathbb{E}[D_{KL}(Q(\cdot|\mathbf{o}_i)\|\mathcal{N}(0,I))].
\end{equation}

\begin{figure*}[t]
\begin{center}
   \vspace{-10pt}
    \includegraphics[width=1\linewidth]{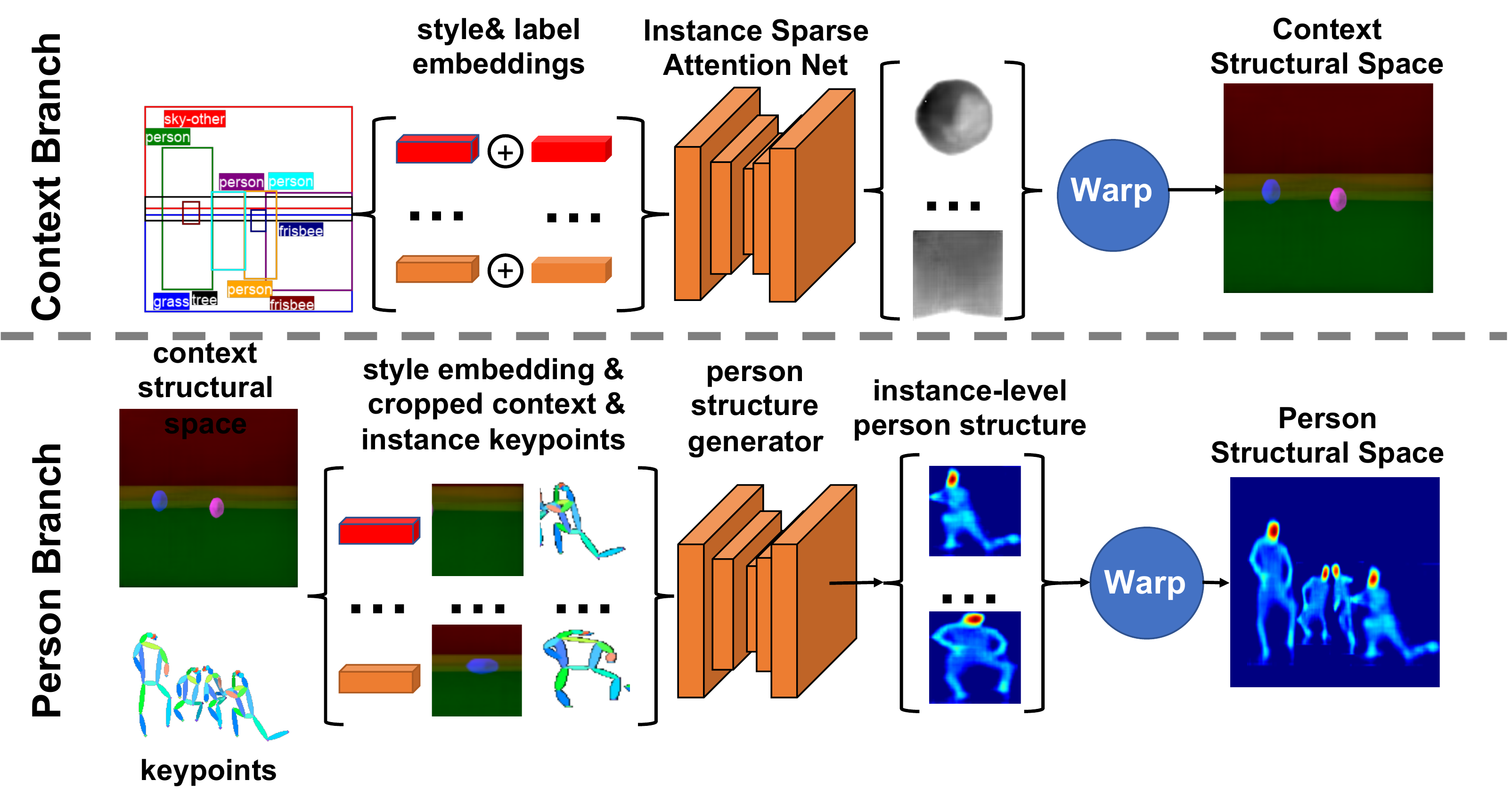}
   \vspace{-20pt}
\end{center}
    \caption{{\bf Illustration of the Two-branches.} Detailed view of context branch and person branch respectively. The input to context branch are label and style embeddings for different instances. Then instance-level sparse attention mask is generated and filled with corresponding embeddings, named as instance-level context structure. The inputs to person branch are instance-level keypoints, style embeddings and cropped context structure. These inputs are converted into instance-level person structure. All instance-level structures are put into locations specified by bounding boxes using differentiable bilinear warpping.}
   \vspace{-20pt}
    \label{fig:two_branch}
\end{figure*}

%\textcolor{red}{1. where ... 2. use equation}

%\begin{figure*}[t]
%\begin{center}
%   \includegraphics[width=1\linewidth]{figure/vg_compare.pdf}
%\end{center}
%\vspace{-15pt}
%   \caption{Examples of generated images from complex layouts on Visual Genome dataset by our proposed method and baselines. For each example we show the input layout and keypoints, ground truth, 64 $\times$ 64 images generated by Layout2im\cite{Zhao2018ImageGF}, 128 $\times$ 128 images generated by LostGAN~\cite{sun2019image}, Scene Generation~\cite{ashual2019specifying} and 256 $\times$ 256 images by our method. }
%\vspace{-15pt}
%\label{fig:vg_compare}
%\end{figure*}

\noindent
\textbf{Location Retargeting by Bilinear Warping.} 
To put different instance-level structures into locations specified by bounding boxes $B$ in a fully differentiable manner, we used differentiable bilinear warpping. This module is shared by person branch and context branch. Given an instance-level structure $\mathbf{f}_i$ with shape $ D \times S_f \times S_f$ and the location specified by $\mathbf{b}_i=\{x_1^i,y_1^i,x_2^i,y_2^i\}$, the warped output $\mathbf{F}_i$ is of size $ D \times S_F \times S_F$ (note $S_f < S_F$). At each spatial location $\mathbf{F}_i(x,y)$, the output feature vector is calculated as 
\begin{equation}
    \mathbf{F}_i(x,y) = \sum_{(x',y') \in N_i(x,y)}(1-|\alpha_x^i x + \beta_x^i - x'|)(1-|\alpha_y^i y+\beta_y^i - y'|) \mathbf{f}_i(x',y')
\end{equation}
where $\alpha_x^i x + \beta_x^i \in (0,S_f), \alpha_y^i y+\beta_y^i \in (0,S_f)$ and $\alpha_x^i=\frac{S_f}{x_2^i-x_1^i}, \beta_x^i=\frac{S_f x_1^i}{x_1^i-x_2^i}, \alpha_y^i = \frac{S_f}{y_2^i-y_1^i},\beta_y^i=\frac{S_f y_1^i}{y_1^i-y_2^i}$. $N_i(x,y)$ denotes the four neighbors of $(\alpha_x^i x+\beta_x^i,\alpha_y^i y+\beta_y^i)$ in $\mathbf{f}_i$. For other locations of $(x,y)$ we simply pad with zeros. 

After bilinear warping of $M$ instance-level structures, we get a tensor $\mathbf{F}$ of shape $M \times D \times S_F \times S_F.$ Then we sum along the first dimension to compose these features together, resulting in the structural space of shape $D\times S_F \times S_F$.

\noindent
\textbf{Context Branch.}
The inputs to context branch are style embeddings $\mathbf{e}_{oi}$ with corresponding label embeddings $\mathbf{e}_{ci}$ for each bounding box $\mathbf{b}_{oi}$ that do not contain person. As is shown in Figure~\ref{fig:two_branch}, instead of filling each bounding boxes with $[\mathbf{e}_{oi},\mathbf{e}_{ci}]$, we first generate an instance-level sparse attention mask for each context object $\mathbf{m}_i=\max(0,G_m(\mathbf{e}_{ci}))$ using a mask generator $G_m$. Given $M = n_o$ objects, the attention masks $M_a=\{\mathbf{m}_1,\dots,\mathbf{m}_{n_o}\}$ are of shape $M \times S_f \times S_f$ where $S_f$ is spatial size of each mask. Then we fill them with embeddings $E=\{[\mathbf{e}_{o1},\mathbf{e}_{c1}],\dots,[\mathbf{e}_{on_o},\mathbf{e}_{cn_o}]\}$ of shape $M \times D$ by cross product and the outputs are $M = n_o$ instance-level structures each of shape $ D \times S_f \times S_f$. Then we use bilinear warping module to put them into correct locations and the output forms context structural space, which is of shape $D\times S_F \times S_F$.

\noindent
\textbf{Person Branch.}
Given $M = n_p$ (with slight abuse of notation) bounding boxes of person $B_p=\{\mathbf{b}_{p1},\dots,\mathbf{b}_{pn_p}\}$ with corresponding keypoints $K=\{\mathbf{k}_1,\dots,\mathbf{k}_{n_p}\}$ inside each box, our goal is to construct person structural space from these inputs similar to that in context branch.
To achieve this goal, we first convert the keypoints $K$ into pose heatmaps $H=\{\mathbf{h}_1,\dots,\mathbf{h}_{n_p}\}$ with size $M \times S_f \times S_f$. The keypoint at each location goes through Gaussian filter with small sigma. To make persons compatible with given context, we also crop out context structures at locations $B_p$ for different persons. Shown in Figure~\ref{fig:two_branch}, given pose heatmaps, cropped context structures and style embeddings for each person, we concatenate them together and introduce a neural person structure generator to get converted person representation $C_p$ of shape $M\times D \times S_f \times S_f$ and sparse attention masks for every person as $M_p$ of shape $M\times 1 \times S_f \times S_f$. Instance-level person structure is constructed as $C'_p=C_p\times M_p$.
%to project them onto instance level person semantic spaces $C_p$. 
Given $C'_p$ of shape $M \times D \times S_f \times S_f$ and bounding boxes $B_p$, we use the same bilinear warping module to put them into correct locations, and the constructed person structural space is of shape $D\times S_F \times S_F$.

The person and context structural spaces from two branches are merged into \textit{compositional structural space} with simple linear summation.
\vspace{-10pt}

\subsection{Image Synthesis from Compositional Structural Space}
\noindent
\textbf{Multi-Level Feature Modulation.}
We get \textit{compositional structural space} $\mathbf{I}_{s}$ from two neural branches. Then we perform \textit{multi-level feature modulation} to convert the structural space into image space.
%Inspired by SPADE~\cite{Park2019GauGANSI} which used segmentation map as the guidance for spatial information, we use the structural space as spatial guidance. 
Specifically, given $\mathbf{I}_{s}$ of shape $D \times S_o \times S_o$, we downsample it into multiple different scales $\{\mathbf{I}_{s}^{S_1},\dots,\mathbf{I}_{s}^{S_n}\}$. At each scale $S_i$ the output from previous module first goes through BatchNorm to obtain output $\mathbf{F}_i$. Then we denormalize $\mathbf{F}_i$: %  as follows:
\begin{equation}
\mathbf{F}'_i=\gamma_i(\mathbf{I}_{s}^{S_i}) * \mathbf{F}_i + \mu_i(\mathbf{I}_{s}^{S_i})
\end{equation}
using two convolutional layers $\gamma_i$ and $\mu_i$ which takes $S_i$ as input. Then the denormalized output is fed into next Residual block as input. Thus the final image is synthesized as $\mathbf{I}'=G_{img}(\mathbf{I}_{s})$.

%Our multi-level feature modulation for compositional structural space is advantageous over the original SPADE, in which the input segmentation mask is fixed during training. In our work this structural space is introduced as an end-to-end trainable component which can be jointly learned by the two separate neural branches to provide finer guidance towards realistic image synthesis.

\vspace{0.05in}
\noindent
\textbf{Person-Context Discriminators.} The realistic output images are generated by jointly training the two neural branches and feature modulation parameters against two discriminators $D_{cxt}$ and $D_{person}$. $D_{cxt}$ operates on the whole image while $D_{person}$ operates on cropped person image patches to provide more training signal for person branch. We used the same patch-based discriminator as pix2pixHD\cite{Wang2017HighResolutionIS} at three different scales. The adversarial loss $\mathcal{L}_{GAN}$ for two discriminators are both calculated as 
\begin{equation}
\mathcal{L}_{GAN}=\mathbb{E`}_{\mathbf{I}\sim p_{\text{real}}}\log D(\mathbf{I}) + \mathbb{E}_{\mathbf{I}'\sim p_{\text{fake}}}\log[1-D(\mathbf{I}')]
\end{equation}
\vspace{-25pt}
\subsection{Learning}

\noindent
\textbf{Training Objectives.} We jointly train the two branches, feature modulation parameters $G_{img}$ and the discriminators $D_{cxt}$, $D_{person}$. The generation network is trained to minimize the weighted sum of following losses:
\vspace{-5pt}
\begin{enumerate}
    \setlength{\itemsep}{0pt}
    \setlength{\parskip}{0pt}
    \setlength{\parsep}{0pt}
    \item  \textit{Feature matching loss: $\mathcal{L}_{feat}=\|F(\mathbf{I}')-F(\mathbf{I})\|$} penalizing the L1 difference between feature vectors of generated images and real images. The features are extracted from discriminator and VGG network.
    
    \item  \textit{KL divergence loss:} $\mathcal{L}_{KL}$ penalizing the KL divergence of posterior distribution $Q(\cdot|\mathbf{o}_i)$ obtained from object embedding network and the normal distribution $\mathcal{N}(0,I)$ prior.
    
    \item  \textit{Image adversarial loss:} $\mathcal{L}_{GAN}$ from discriminator encouraging the generated image patches to appear realistic. We use a hinge loss, which is a variant of GAN loss.
    
    \item  \textit{Attention TV loss: $\mathcal{L}_{attn} = \sum_i \norm{\nabla \Phi^{x_i}}^2 + \norm{\nabla \Phi^{y_i}}^2$} on instance level sparse attention $\mathbf{m}_i$ both for person and context to regularize the attention mask to be smooth with fewer holes.
\end{enumerate}
\noindent
\textbf{Implementation Details.} We train all models using Adam with learning rate $2\times 10^{-4}$ for 100 epochs both on COCO and Visual Genome dataset. We use batch size 8 for each GPU at 256 resolution and 32 at 128 resolution. We use 4 Tesla P100 in parallel and the model converges in 5 days at 256 and 1 day at 128 resolution. We use LeakyReLU for both generator and discriminator. % in our experiments. 
%For details please see supplementary materials; {\bf we will also release code}.
\vspace{-5pt}

\vspace{-10pt}
\section{Experiments}
We evaluated our model at two different resolutions on Visual Genome and COCO-Stuff datasets. In our experiments we aim to show that our method generates images of complex layouts which respect the input bounding boxes, class labels and keypoints. As there's no existing methods that specifies both layouts and keypoints as input, we divide our comparison into two sections. In the first section we compare with all standard baselines. In the second section we compare with state-of-the-art variants and ablations that specify both layout and person annotation as input for a detailed analysis. We will release the code upon acceptance.

\begin{figure*}[t]
\begin{center}
    \vspace{-10pt}
   \includegraphics[width=1\linewidth]{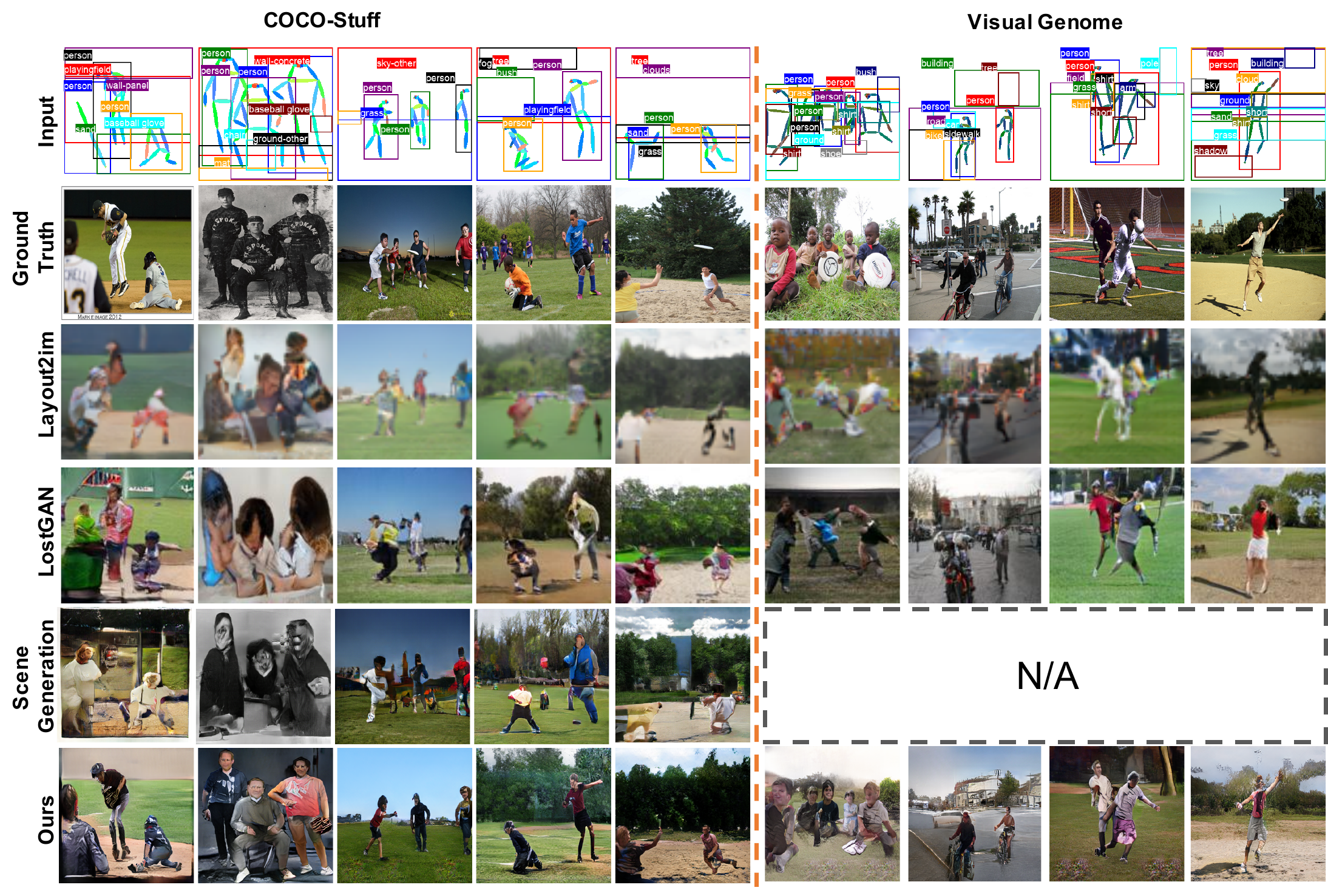}
\end{center}
\vspace{-10pt}
   \caption{{\bf Examples of generated images from complex layouts.} Results on COCO-Stuff and Visual Genome obtained by our method and the baselines. For each example we show input layout with keypoints, ground truth, 64 $\times$ 64 images generated by Layout2im~\cite{zhao2019image}, 128 $\times$ 128 images generated by LostGAN~\cite{sun2019image} and 256 $\times$ 256 images by Scene Generation~\cite{ashual2019specifying} and our method. Note that ~\cite{ashual2019specifying} only have results on COCO-Stuff.}
\vspace{-15pt}
\label{fig:all_compare}
\end{figure*}

% Please add the following required packages to your document preamble:
% \usepackage{booktabs}
% \usepackage{multirow}
\begin{table}[t]
\caption{A quantitative comparison using various image generation scores on person split of COCO-Stuff and Visual Genome dataset.}
\resizebox{\textwidth}{!}{
% Please add the following required packages to your document preamble:
% \usepackage{booktabs}
% \usepackage{multirow}

\begin{tabular}{@{}cccccccccccc@{}}
\toprule
\multicolumn{2}{c}{Datasets}                & \multicolumn{4}{c}{COCO-Stuff}          & \multicolumn{4}{c}{Visual Genome}      & \multicolumn{2}{c}{Param Num} \\ \midrule
Resolution               & Method           & IS         & FID    & Acc   & DS        & Inception  & FID   & Acc   & DS        & G              & D            \\ \midrule
\multirow{4}{*}{128x128} & Real Im      & 17.30$\pm$0.14 & 0.00   & 58.51 & -         & 17.41$\pm$0.16 & 0.00  & 63.24 & -         & -              & -            \\
                         & SG\cite{ashual2019specifying} & 9.17$\pm$0.66  & 85.83  & 39.92 & 0.35$\pm$0.08 & -          & -     & -     & -         & 183.07         & \textbf{1.50}         \\
                         & LostGAN\cite{sun2019image}          & \textbf{9.35$\pm$0.52}  & 78.20  & 41.10 & \textbf{0.40$\pm$0.09} & \textbf{8.26$\pm$0.35}  & 62.10 & 40.94 & \textbf{0.43$\pm$0.09} & 36.30          & 57.88        \\
                         & Ours             & 8.95$\pm$0.15  & \textbf{77.80}  & \textbf{50.17} & 0.33$\pm$0.12 & 7.68$\pm$0.46  & \textbf{58.74} & \textbf{57.49} & 0.32$\pm$0.09 & \textbf{22.70}          & 4.40         \\ \midrule
\multirow{3}{*}{256x256} & Real Im      & 20.22$\pm$0.77 & 0.00   & 61.77 & -         & 22.63$\pm$0.23 & 0.00  & 65.82 & -         & -              & -            \\
                         & SG\cite{ashual2019specifying} & 10.33$\pm$0.43 & 103.80 & 37.84 & \textbf{0.48$\pm$0.09} & -          & -     & -     & -         & 183.07         & \textbf{1.50}         \\
                         & Ours             & \textbf{10.92$\pm$0.41}& \textbf{76.10}  & \textbf{51.08} & 0.38$\pm$0.09 & 10.61$\pm$0.43 & \textbf{60.86} & \textbf{58.88} & \textbf{0.36$\pm$0.10} & \textbf{35.10}          & 4.40         \\ \bottomrule
\end{tabular}
}
\vspace{-20pt}
\label{tab:all_comparision}
\end{table}

\vspace{-10pt}
\subsection{Benchmark Results}
\noindent
\textbf{Datasets.} We perform experiments on the 2017 COCO-Stuff \cite{caesar2018cvpr} dataset, which augments a subset of the COCO dataset with additional stuff categories. The dataset annotates 40K train and 5K validation images with bounding boxes for 182 categories in total. 

We set the maximum number of bounding boxes to appear in one image as 12. In practice, we sort the bounding boxes in a descending order of area and keep the top 12 bounding boxes with largest area, removing the rest.
We also remove images with objects covering less than $70\%$ of the area, and those without any bounding boxes containing keypoints, leaving around $55K$ images for training.  To evaluate the performance of all models under person-in-context setting, we remove images in the validation/test set that do not contain any person. We name it as ``person split" for COCO which gives us around 1K images. We will release the corresponding splits. %To make a fair comparison, we use the same validation set as \cite{johnson2018image} in Table~\ref{tab:all_comparision}.

We also used Visual Genome \cite{krishna2017visual} version 1.4 which comprises around $110K$ images annotated with bounding boxes. We divide the data into $80\%$ train, $10\%$ val and $10\%$ test using same splits as \cite{johnson2018image}. Also we use the same label set as \cite{johnson2018image}, except that we use one label `person' for all instances of `woman', `man', etc. %We remove small bounding boxes and images with little object coverage following the same procedure as for COCO. 
Finally, we use AlphaPose \cite{fang2017rmpe,xiu2018poseflow} to detect keypoints automatically in all images. The original split gives us around $60K$ images for training and $5K$ for testing. And similarly, we evaluate on the ``person split" of Visual Genome which contains around 2K images. %We keep images that contain no people for training. % person in training set. 
% If no person is present inside an image we just train the context branch.

\begin{figure}[t]
\begin{center}
   \includegraphics[width=\linewidth]{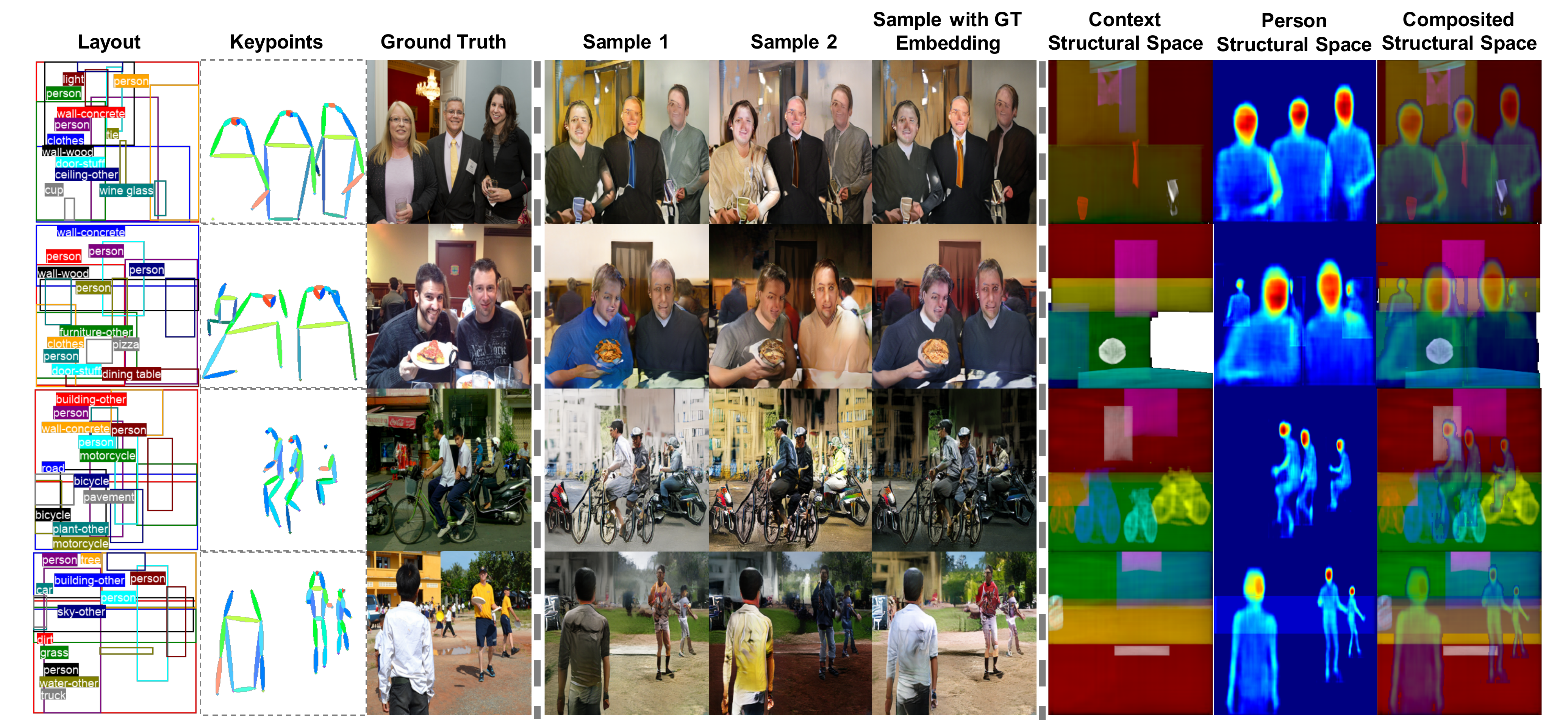}
\end{center}
\vspace{-10pt}
   \caption{Examples of generated images from different style embeddings and the corresponding visualized structural space. The first three columns are ground truth layouts, keypoints and images. The next three columns are synthesized images at 256 resolution from different style embeddings(two randomly sampled and one extracted from ground truth images). The last three columns are visualized context, person and compositional structural space respectively.}
\vspace{-25pt}
\label{fig:coco_diversity}
\end{figure}

\begin{figure}[t]
\begin{center}
   \includegraphics[width=\linewidth]{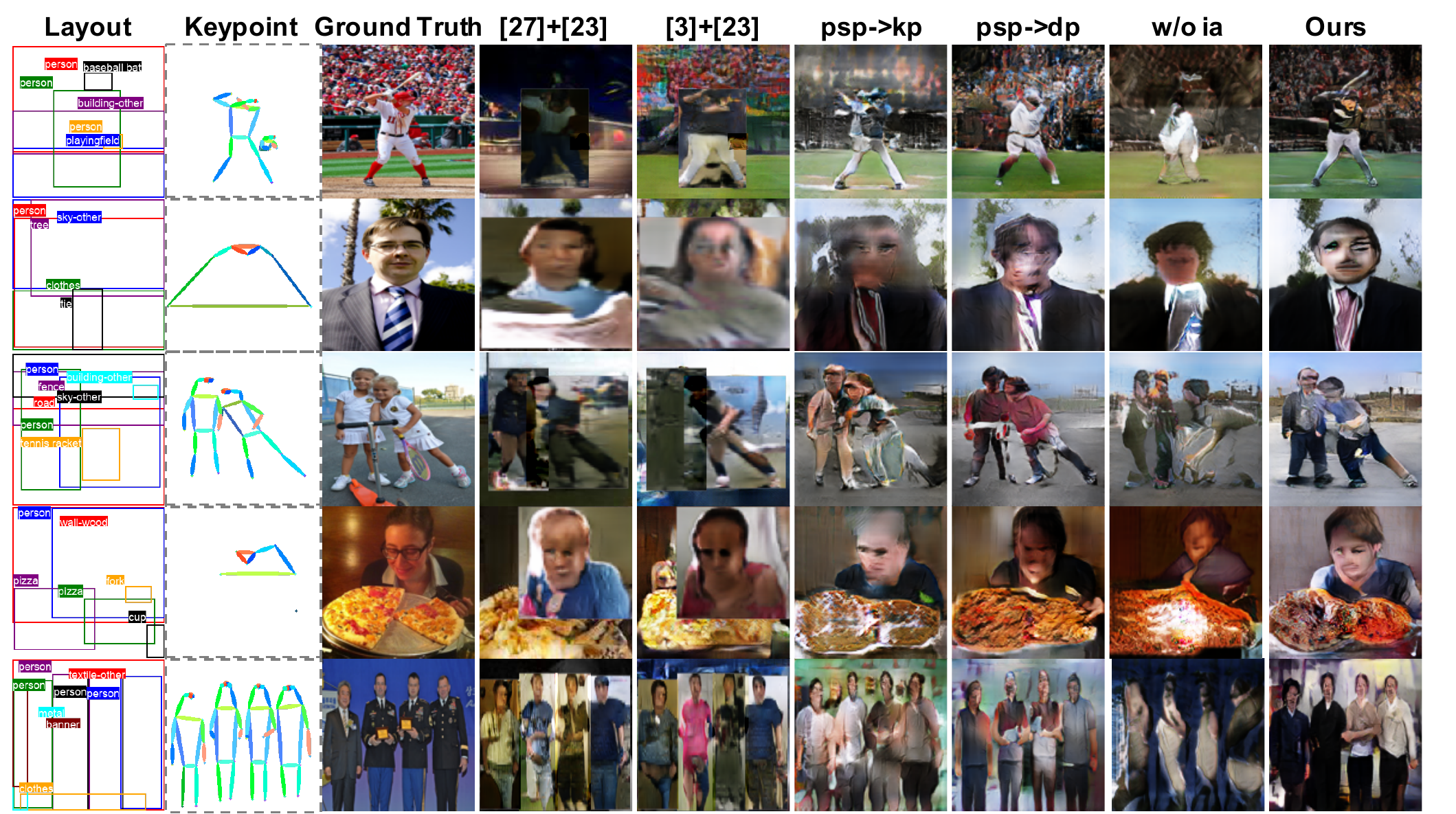}
\end{center}
   \vspace{-5pt}
   \caption{Examples of generated images from state-of-the-art variants and ablations using both layout and person annotations. For each example we show input layouts, input images, ground truth, two baselines from existing methods, three ablations and our method on COCO-Stuff dataset at 128 resolution.}
   \vspace{-20pt}
\label{fig:fair_comparison}
\end{figure}

\noindent
\textbf{Standard Comparison Methods.}
We compare our approach with several existing state-of-the-art image synthesis methods. %Sg2im \cite{johnson2018image} and 
Scene Generation(SG) \cite{ashual2019specifying} generate images from scene graphs. For fair comparison, we use ground truth layout for them to generate images. The \cite{ashual2019specifying} requires mask annotation so only results on COCO-Stuff are available for this method. LostGAN \cite{sun2019image} generate images directly from given layout. As different methods work under different resolutions, we report results for two different resolutions at 128$\times$128 and 256$\times$256. Sg2im\cite{johnson2018image} and Layout2Im\cite{zhao2019image} only works under $64\times64$ resolution so we did not compare with those quantitatively.

%
%GauGAN\cite{Park2019GauGANSI} is originally designed for image synthesis from segmentation mask. For fair comparison We convert layouts to segmentation masks where each instance is a rectangle.

%\vspace{0.05in}
\noindent
\textbf{Evaluation Metrics.} We adopt multiple evaluation metrics for evaluating the generated images. Frechet Inception Distance (FID) \cite{heusel2017gans} is employed to measure the distribution distance between generated images and real images. The lower the better. Diversity Score(DS) \cite{zhang2018perceptual} is used to measure the distance between pairs of images generated given same input. It is based on the perceptual similarity between two images. The higher the better. Inception Score (IS) \cite{salimans2016improved} is also used to evaluate the quality of generated images. It uses an ImageNet classification model to encourage recognizable objects within images and diversity across images. Classification Accuracy (Acc) is used to evaluate whether the generated objects are recognizable. The higher the better. We trained a ResNet50 classifier on real images with two different scales to serve as an oracle.

\noindent
\textbf{Qualitative Results.}
Figure~\ref{fig:all_compare} shows generated images using our method as well as the baselines. As can be seen we can generate complex images with multiple objects at high resolution and with realistic details. For example, in column two our method generates three persons with diverse textures, and different parts of the person are recognizable, such as heads, hands, legs and shoes. The other methods failed to produce recognizable person 
appearances. 
%due to the nature of complex structures and diversities inside persons. 
These examples also show that our method generates images which respect the location constraint, class constraint and keypoints constraint. This is due to the superiority of our combination of compositional structural space and feature modulation techniques, which projects annotations with different modalities into shared structural space such that they are compatible during generation process.

\noindent
\textbf{Diverse Sampling from Style Embeddings.} In Figure~\ref{fig:coco_diversity} we demonstrate our method's ability to generate a diverse set of images given the same layout, by sampling from different style codes which follow Gaussian prior. Since we used VAE to construct the latent space of style codes, we can easily manipulate the style of different objects by providing different style codes. For example, in column ``Sample 1" and ``Sample 2", the sampled style embeddings from Gaussian prior are  completely different from each other. And the ``Sample with GT Embedding" column use embeddings extracted from ground truth images, resulting in output images that possess similar appearances as ground truth while maintaining same structures. This disentanglement is enabled by compositional structural space.
% thus the appearance is controlled and looks similar to ground truth images.

\noindent
\textbf{Quantitative Results.} 
Table~\ref{tab:all_comparision} compares our method with other baselines and the real test images using person splits on COCO-Stuff and Visual Genome. %from \cite{johnson2018image}. (Refer to the supplementary material for more details.) As~\cite{ashual2019specifying} needs ground truth masks for training, results on Visual Genome are not available. % We did not use any information from the test images by sampling style codes from Gaussian distribution. 
%Our proposed method outperforms other methods in terms of FID and Classification accuracy. This is because our method is advantageous when resolution goes up, where the structure of different objects and persons become more important and directly influences the final performance. Our diversity score is not as good as some of the other baselines. This is because our method respects the input specified by compositional structural space, and the diversity sampling will only change the texture of generated images instead of the structure as is shown in Figure~\ref{fig:coco_diversity}. We also achieved highest Inception score at 256 resolution. 
Our method outperforms other method in terms of FID and Classification Accuracy. 
We noticed that LostGAN achieved comparable performance as our model, and even better in terms of Inception Score. This is due to their discriminator which has an order of magnitude higher number of parameters. As is shown in Table~\ref{tab:all_comparision}, their discriminator has 57.88 millon parameters, which does not scale up to higher resolutions. Instead, SG and our work borrow discriminator from patchGAN which requires significantly less parameters (1.5 and 4.4 million respectively). As a result, our method is more stable during training, requires less computational cost and scales to higher resolutions. With the same patchGAN based discriminator, our method beats SG by a large margin. Our diversity score is not as good as some of the other baselines. This is because our method respects the input specified by compositional structural space, and the diversity sampling will only change the texture of generated images instead of the structure as is shown in Figure~\ref{fig:coco_diversity}.

%There is some inconsistencies between FID score and Inception Score. We argue that FID score is more reliable than Inception Score, which is verified in~\cite{heusel2017gans}. 
\vspace{-10pt}

\begin{figure}[t]
\begin{center}
   \includegraphics[width=\linewidth]{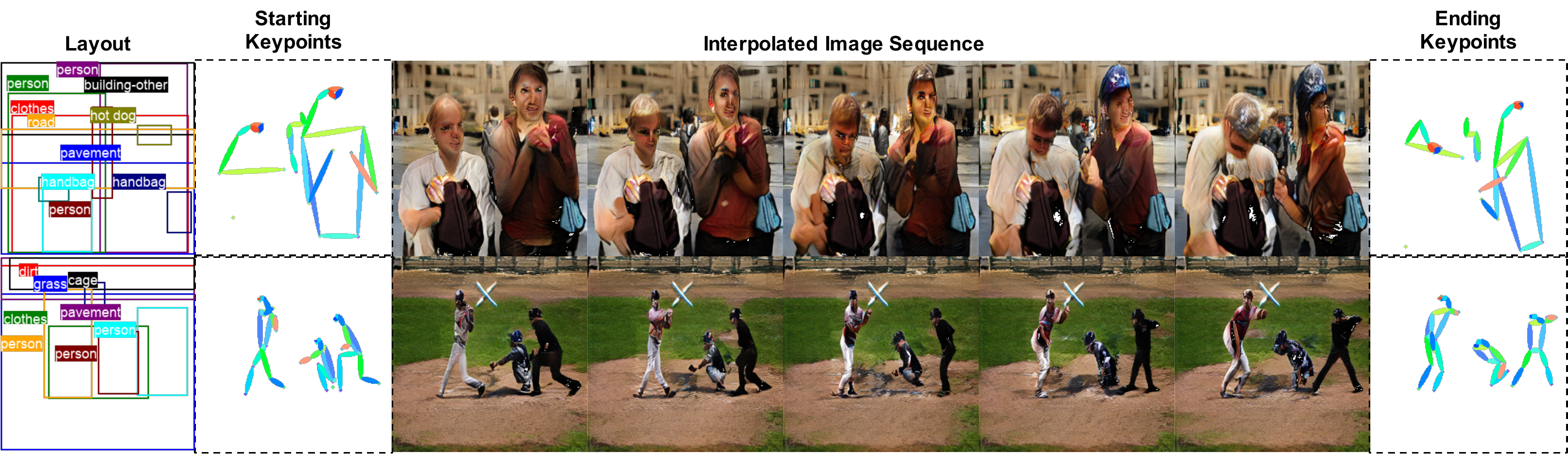}
\end{center}
   \vspace{-5pt}
   \caption{Examples of generated images by interpolating poses while keeping appearance the same. For each example we show layouts, starting poses, interpolated image sequence, and ending poses on COCO-Stuff dataset at 256 resolution.}
   \vspace{-20pt}
\label{fig:interp_pose}
\end{figure}
\subsection{Comparison with State-of-the-Art Variants and Ablations}
\textbf{State-of-the-Art Variants.}
There is no existing method that addresses the problem of person in context synthesis, which specifies both layout and keypoint as input. Thus we proposed several variants, which require both layout annotations and person annotations such as keypoints or densepose masks\cite{alp2018densepose}. Two variants(\cite{Park2019GauGANSI}+\cite{sun2019image},\cite{Park2019GauGANSI}+\cite{ashual2019specifying}) are proposed based on existing state-of-the-art. GauGAN \cite{Park2019GauGANSI} specifies one pose heatmap as input and synthesizes one single person each time. We trained it from scratch for keypoint guided pose synthesis. Then we combine results with \cite{sun2019image} and \cite{ashual2019specifying}, respectively, by blending the synthesized person image patches with synthesized images from layout at corresponding person box locations using Poisson blending.

We also demonstrate that a naive combination of context and pose annotations does not succeed, neither for sparse keypoints nor dense segmentation masks, by providing three ablations that take both of these annotations. ``psp$\rightarrow$kp" replaces person structural space with keypoints, which is concatenated directly on top of context structural space. Similarly, ``psp$\rightarrow$dp" replaces person structural space with densepose masks, which is a series of 2d segmentation masks that annotates the shape of different body parts. Densepose masks are available on COCO dataset. Note that these masks are more powerful and expensive annotations as compared with 2d keypoints used by us. ``w/o ia" removes the instance-level sparse attention during construction process of compositional structural space. % All these ablations are trained from scratch.

\noindent
\textbf{Person Crop Datasets.} 
To evaluate the synthesize quality of person images, we construct another dataset named `person crop'. It is constructed from COCO images and each person crop is resized into 64$\times$64 patch. The training and testing split for person crop is same as COCO. We use the training split for GauGAN to learn from scratch, and the testing split to evaluate different methods.  To compare with, we crop out persons from generated images at 128 resolution and resize them into $64\times 64$ patches. Results are in Table~\ref{tab:ps_pc}.

\noindent
\textbf{Effectiveness of Compositional Structural Space.} As is shown in Figure~\ref{fig:fair_comparison}, the boundary between person and context looks seamless in our method (the last column), while the blended person(s) look unnatural for \cite{ashual2019specifying}+\cite{Park2019GauGANSI} and \cite{sun2019image}+\cite{Park2019GauGANSI}. This is also validated in Table~\ref{tab:ps_pc}, where our method achieves the lowest FID on both `person split' and `person crop'. If we look at the performance difference between \cite{sun2019image} and \cite{sun2019image}+\cite{Park2019GauGANSI}, or \cite{ashual2019specifying} and \cite{ashual2019specifying}+\cite{Park2019GauGANSI}, there is a performance drop with \cite{Park2019GauGANSI} added. This leads to the conclusion that modeling layouts and keypoints separately in image space will decrease the performance after blending. By projecting them into the same compositional structural space, we get more coherent and compatible results when it is decoded into an image.

\noindent
\textbf{Person Reenactment under Context.} In Figure~\ref{fig:interp_pose} we reenact the persons in synthesized context by interpolating between starting keypoints and ending keypoints, while keeping style embeddings fixed for both the person and context. Shown in the first row, orientation of faces of each person is changed gradually. And in the second row, overall body structures are interpolated smoothly. This is enabled by the compositional structural space, which is a disentangled representation of person and context structures.

\noindent
\textbf{Effectiveness of Person Structural Space}. Compared with results of ``psp$\rightarrow$kp" and ``psp$\rightarrow$dp" shown in Figure~\ref{fig:fair_comparison} (6th and 7th column), our method shows more clear body parts and higher quality context. For example, in the 2nd row the face of synthesized person looks more clear, and in the 3rd row the context looks more compatible. Further, as is shown in Table~\ref{tab:ps_pc}, our method achieves the lowest FID and highest IS score compared with these ablations. This validates the conclusion that stronger annotations (such as densepose mask) does not necessarily produce higher quality results. Person keypoint annotations lie in a different structural spaces from context. Naive concatenation of keypoints on top of context structural space leads to performance drop.
% during feature modulation. 

We visualized person structural spaces with heatmaps using L1 norm of corresponding feature vectors in Figure~\ref{fig:coco_diversity}. The visualized person features are dense around relevant body parts, highly activated around head (shown in red) and joints (shown in green) and not activated in irrelevant regions. This learned representation has richer structures than raw annotations such as keypoints or densepose masks, and are more compatible with context representations.

% Please add the following required packages to your document preamble:
% \usepackage{booktabs}
% \usepackage{multirow}
\begin{table}[t]
\caption{Qualitative results on proposed person split and person crop dataset. $^1$ only use layout as input. $^2$ use both layout and keypoint. $^3$ use both layout and densepose mask.}
\resizebox{\textwidth}{!}{
\begin{tabular}{@{}ccccccccccc@{}}
\toprule
\multicolumn{2}{c}{Method}                & \cite{sun2019image}$^1$        & \cite{ashual2019specifying}$^1$        & \cite{sun2019image}+\cite{Park2019GauGANSI}$^2$ & \cite{ashual2019specifying}+\cite{Park2019GauGANSI}$^2$ & psp$\rightarrow$kp$^2$ & psp$\rightarrow$dp$^3$ & w/o ia $^2$    & ours$^2$      & ground truth \\ \midrule
\multirow{2}{*}{Person Split } & FID$\downarrow$       & 78.20      & 85.83     & 98.07    & 100.27    & 99.75               & 100.43              & 94.74     & \textbf{77.80}     & 0            \\
                              & IS$\uparrow$ & \textbf{9.35$\pm$0.52} & 7.39$\pm$0.27 & 7.08$\pm$0.37 & 6.03$\pm$0.34 & 6.52$\pm$0.34           & 7.53$\pm$0.51           & 7.50$\pm$0.04 & 8.95$\pm$0.15 & 17.00$\pm$0.28   \\ \midrule
\multirow{2}{*}{Person Crop }  & FID$\downarrow$       & 80.60      & 81.44     & 86.84     & 86.84     & 77.74               & 75.26               & 77.57     & \textbf{52.81}     & 0            \\
                              & IS$\uparrow$ & 5.82$\pm$0.19 & 5.99$\pm$0.10 & 4.09$\pm$0.06 & 4.09$\pm$0.06 & 6.01$\pm$0.13           & 5.77$\pm$0.03           & 5.92$\pm$0.05 & \textbf{6.19$\pm$0.25} & 7.92$\pm$0.35    \\ \bottomrule
\end{tabular}
}
\label{tab:ps_pc}
\vspace{-20pt}
\end{table}
% Please add the following required packages to your document preamble:
% \usepackage{booktabs}
\begin{table}[b]
\vspace{-25pt}
\centering
\caption{User Study Results on COCO-Stuff Dataset at 128$\times$128 resolution.}
\begin{tabular}{@{}lcc@{}}
\toprule
Method     & \begin{tabular}[c]{@{}c@{}}Global \\ Coherence\end{tabular} & \begin{tabular}[c]{@{}c@{}}Visual Quality\\  of Persons\end{tabular} \\ \midrule
LostGAN~\cite{sun2019image}    & 35\%                                                        & 15\%                                                                 \\
Scene Generation~\cite{ashual2019specifying} & 20\%                                                        & 10\%                                                                 \\
Ours       & 45\%                                                        & 75\%                                                                 \\ \bottomrule
\end{tabular}
\label{tab:user_study}
\end{table}
\noindent
\textbf{Effectiveness of Instance Level Sparse Attention.} 
As is shown in Figure~\ref{fig:coco_diversity}, each instance structure (context and person), is zero at irrelevant regions.
%(mostly person and foreground objects). 
As shown in Figure~\ref{fig:fair_comparison} and Table~\ref{tab:ps_pc}, the removal of this sparse attention mask will lead to performance drop, because: 1) different bounding boxes can affect each other in overlapping areas and 2) the shape 
of the instances is less accurate.

\vspace{-10pt}
\subsection{Further Analysis}

\begin{figure}[t]%
    \centering
    %\hspace{-10pt}
    \subfloat[FID score for images with different number of persons. Maximum number of persons is 3.]{{\includegraphics[width=0.45\linewidth]{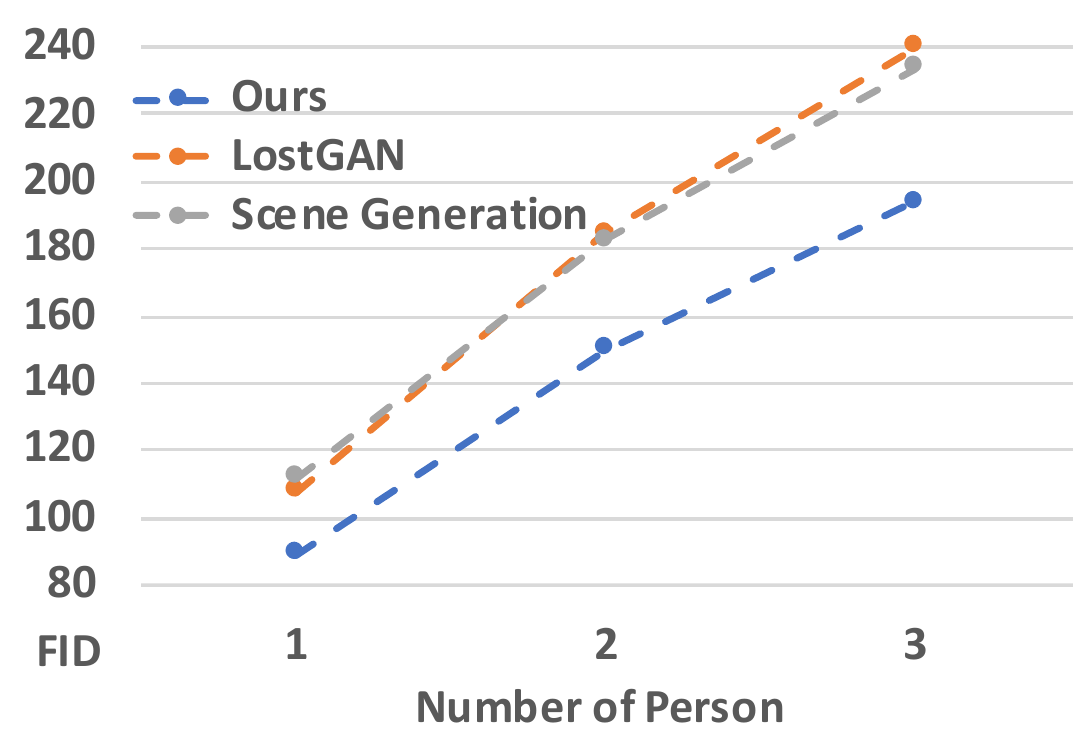} }}%
    \qquad
    \hspace{-0pt}
    \subfloat[FID score on cropped person images divided into different pose clusters. Each cluster has more than 300 images.]{{\includegraphics[width=0.45\linewidth]{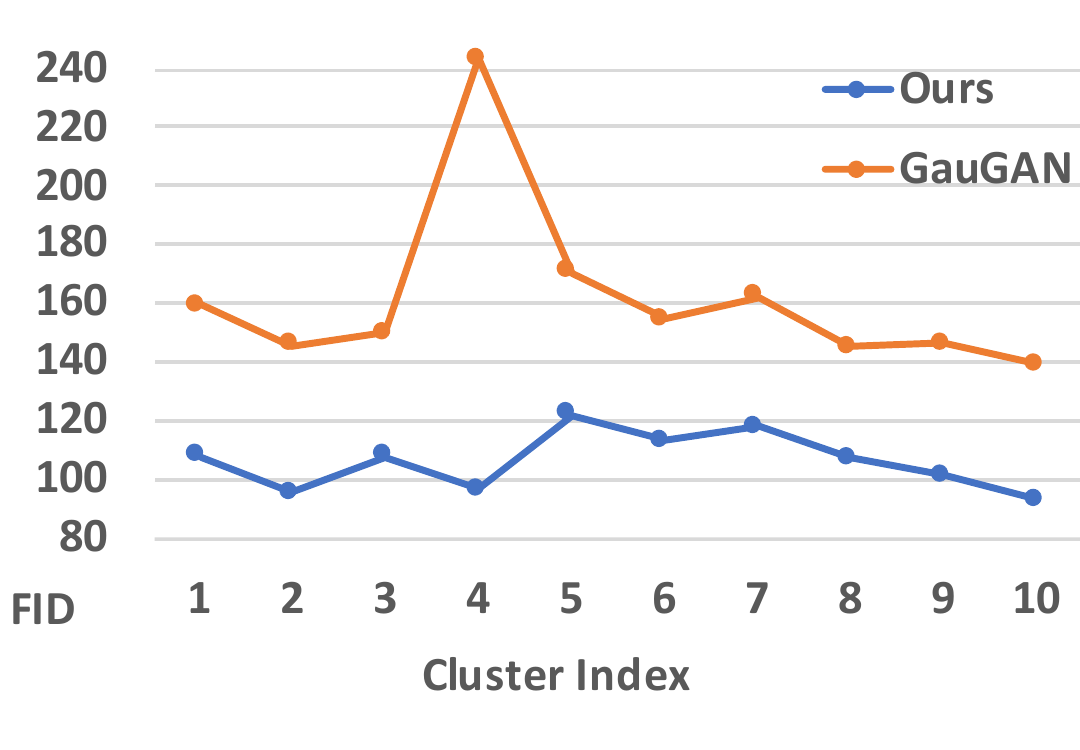} }}%
    %\hspace{-10pt}
    \caption{Performance of different models under complex scenes.}%
    \vspace{-20pt}
    \label{fig:diff_person_cluster_fid}%
\end{figure}

\noindent
\textbf{Performance under Complex Scenes.} We evaluate the performance of our model under scenes with multiple persons and diverse poses. In Figure~\ref{fig:diff_person_cluster_fid}(a), validation sets are divided into three groups with number of persons as criterion. When only one person is present, our model performs slightly better. As number of persons increases, the difference between \cite{ashual2019specifying,sun2019image} and our method becomes more clear. This is because our model can deal with challenging inputs containing multiple persons. Shown in Figure~\ref{fig:diff_person_cluster_fid}(b), we cluster poses and evaluate FID on different clusters. Performance of ~\cite{Park2019GauGANSI} is inconsistent among  clusters, while our model achieves lower FID score and is more consistent for different type of poses. %This validates the ability of compositional structural space to model a diverse set of poses under diverse contexts.

\vspace{0.05in}
\noindent
\textbf{User Study.} We perform a user study to compare with other baselines. 20 volunteers were involved. Each volunteer was shown the synthesized images from COCO-Stuff dataset at 256 resolution and was asked to select the preferable images
in terms of the global coherence of both context and persons, and the visual quality of persons respectively. The results reported in Table~\ref{tab:user_study} show that our method significantly outperforms other methods, especially in terms of visual quality of synthesized persons.

\vspace{-10pt}
\section{Conclusion}
   We proposed a novel problem called \textbf{Persons in Context Synthesis}, which aims to synthesize 1) diverse person instances, as well as 2) varying contexts that are visually compatible with the synthesized persons. The context is specified by %the sparse 
bounding box object layout, while pose of the person(s) by keypoints. %the appearance of both is modeled by per-instance latent code, inferred or drawn from a distribution.   
This difference in input modalities motivate the use of separate neural branches that attentively project the respective (context/person) inputs into ``compositional structural space'', where person and context representations are compatible with each other. %Furthermore, we design multi-level feature modulation strategy and person-context discriminator to learn this semantic space in a weakly supervised manner. 
Extensive experiments on two large-scale datasets (COCO-Stuff and Visual Genome) demonstrate that our approach outperforms state-of-the-art in synthesis quality and diversity.
   
\clearpage
% ---- Bibliography ----
%
% BibTeX users should specify bibliography style 'splncs04'.
% References will then be sorted and formatted in the correct style.
%
\bibliographystyle{splncs04}
\bibliography{egbib}
\end{document}